\documentclass[10pt,twocolumn,letterpaper]{article}

\usepackage[pagenumbers]{cvpr} 

\usepackage[dvipsnames]{xcolor}

\usepackage{graphicx}
\usepackage{amsmath}
\usepackage{amssymb}
\usepackage{booktabs}
\usepackage{bm}
\usepackage{multirow}
\usepackage{pifont}
\usepackage{booktabs}
\usepackage{colortbl}
\usepackage{xspace}
\usepackage{wrapfig}
\usepackage{hhline}
\usepackage{tabularx}
\usepackage{makecell}
\usepackage[ruled,linesnumbered]{algorithm2e}

\usepackage{wrapfig}
\usepackage{colortbl}
\definecolor{cvprblue}{rgb}{0.21,0.49,0.74}
\usepackage[pagebackref,breaklinks,colorlinks,citecolor=cvprblue]{hyperref}
\usepackage{soul}

\definecolor{mypurple}{RGB}{170,120,200}
\definecolor{floor}{RGB}{0,0,255}
\definecolor{chair}{RGB}{255,0,0}
\definecolor{bookcase}{RGB}{10,200,100}
\definecolor{mygray}{gray}{.9}
\definecolor{board}{RGB}{255,192,203}
\newcommand{\matdn}[2]{\mathbf{#1}_{\rm #2}}

\newcommand{\matud}[3]{\mathbf{#1}_{\rm #2}^{\rm #3}}
\newcommand{\myPara}[1]{\vspace{.05in}\noindent\textbf{#1}\quad}

\def\ourmodel{COSeg}
\def\paperID{} 
\def\confName{CVPR}
\def\confYear{2024}

\title{Rethinking Few-shot 3D Point Cloud Semantic Segmentation}

\author{Zhaochong An$^{1,2}$, \quad Guolei Sun$^{1*}$, \quad Yun Liu$^3$\thanks{Corresponding authors: Guolei Sun and Yun Liu}, \quad Fayao Liu$^3$, \quad Zongwei Wu$^4$,\\ 
Dan Wang$^2$, \quad Luc Van Gool$^{1}$, \quad Serge Belongie$^2$\\
$^1$ Computer Vision Laboratory, ETH Zurich\\
$^2$ Pioneer Centre for Artificial Intelligence, University of Copenhagen\\
$^3$ Institute for Infocomm Research, A*STAR\\
$^4$ Computer Vision Lab, CAIDAS \& IFI, University of Wurzburg \\
}

\begin{document}

\maketitle

\begin{abstract}
This paper revisits few-shot 3D point cloud semantic segmentation (FS-PCS), with a focus on two significant issues in the state-of-the-art: foreground leakage and sparse point distribution. The former arises from non-uniform point sampling, allowing models to distinguish the density disparities between foreground and background for easier segmentation. The latter results from sampling only 2,048 points, limiting semantic information and deviating from the real-world practice. To address these issues, we introduce a standardized FS-PCS setting, upon which a new benchmark is built. Moreover, we propose a novel FS-PCS model. While previous methods are based on feature optimization by mainly refining support features to enhance prototypes, our method is based on correlation optimization, referred to as Correlation Optimization Segmentation (\ourmodel). Specifically, we compute Class-specific Multi-prototypical Correlation (CMC) for each query point, representing its correlations to category prototypes. Then, we propose the Hyper Correlation Augmentation (HCA) module to enhance CMC. Furthermore, tackling the inherent property of few-shot training to incur base susceptibility for models, we propose to learn non-parametric prototypes for the base classes during training. The learned base prototypes are used to calibrate correlations for the background class through a Base Prototypes Calibration (BPC) module. Experiments on popular datasets demonstrate the superiority of \ourmodel~over existing methods.
The code is available at \href{https://github.com/ZhaochongAn/COSeg}{github.com/ZhaochongAn/COSeg}.
\end{abstract}

\begin{figure}[t!]
    \centering
    \includegraphics[width=.95\linewidth]{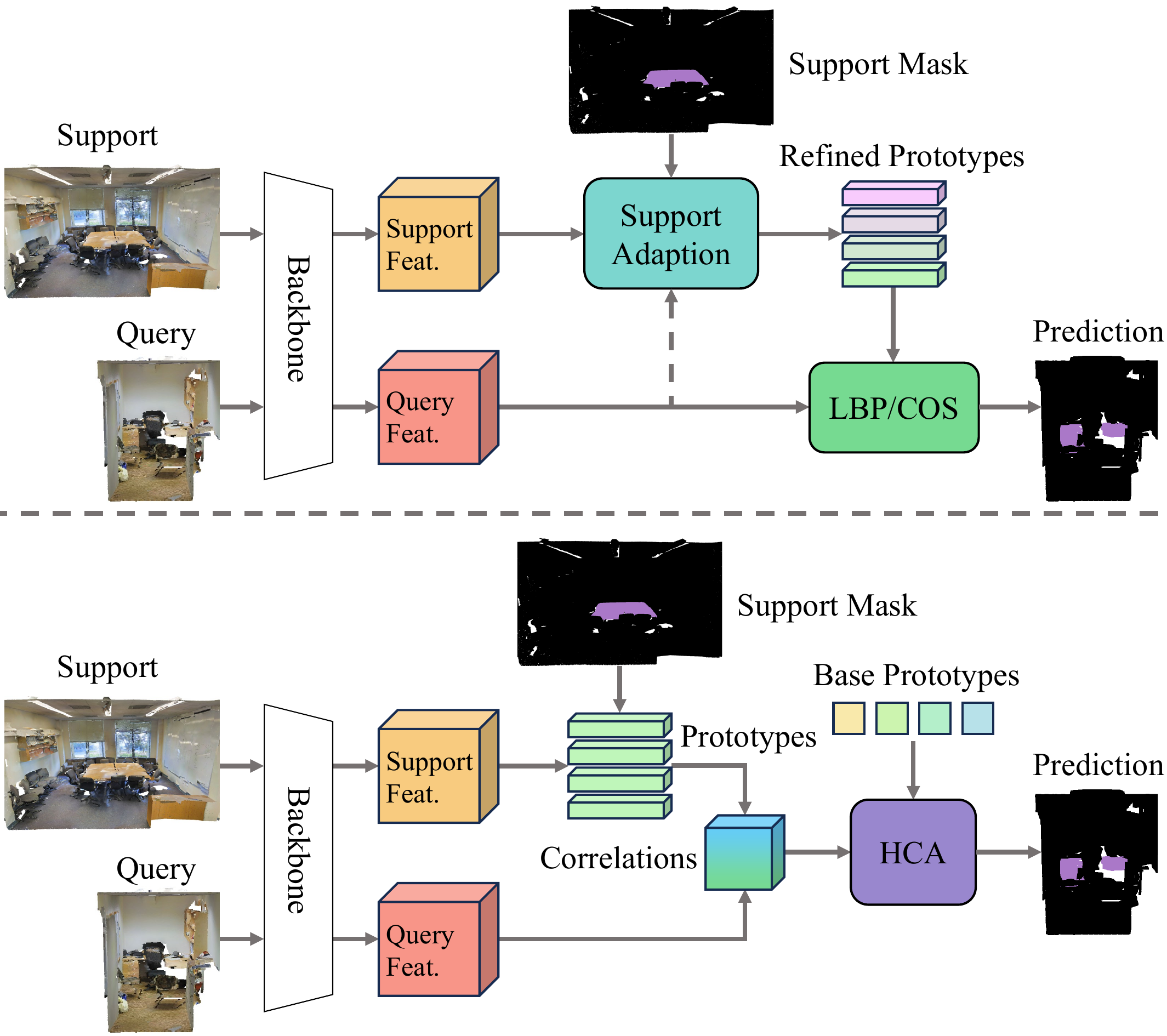}
    \vspace{-.10in}
    \caption{\textbf{Previous \textit{feature optimization} \vs our \textit{correlation optimization}.} 
    \textit{Top:} Most prior work~\cite{zhao2021few,he2023prototype,ning2023boosting,zhu2023cross,mao2022bidirectional,wang2023few} on FS-PCS focuses on feature optimization by designing support adaption modules for enhanced prototypes and then making predictions through non-parametric label propagation (LBP) or cosine similarity (COS), implicitly modeling correlations. 
    \textit{Bottom:} Instead of optimizing features, we propose to directly uses correlations as input to learnable modules, explicitly refining correlations.
    }
    \label{fig:intro}
    \vspace{-.15in}
\end{figure}

\section{Introduction} \label{sec:intro}
Rapid advancements in deep neural networks have propelled the exploration of 3D point cloud understanding in various applications~\cite{engelcke2017vote3deep,chen2017multi,milioto2019rangenet++,pierdicca2020point}.
Unlike images, point clouds inherently capture intricate object structures, enabling fine-grained analyses.
However, collecting and annotating point cloud data is significantly more labor-intensive than its 2D counterpart, limiting the scale and semantic diversity of existing 3D datasets~\cite{behley2019semantickitti,dai2017scannet,armeni20163d}. 
To reduce the substantial human effort required for dataset creation, 
\textit{few-shot point cloud semantic segmentation} (FS-PCS) emerges as a crucial task, which empowers 3D segmentation models to generalize to novel classes with few annotated samples.

In the realm of FS-PCS, attMPTI~\cite{zhao2021few} stands as a pioneering model, introducing a multi-prototype transductive approach that leverages label propagation for predicting segmentation in novel classes.
Subsequent works~\cite{he2023prototype,ning2023boosting,zhu2023cross,mao2022bidirectional,wang2023few,zhang2023few} have consistently built upon the attMPTI framework, progressively improving overall performance.

However, we identify two significant issues in the current FS-PCS setting:
(1) The first issue is the \textit{foreground leakage}. 
The common 3D segmentation practice~\cite{lai2022stratified,zhao2021point} feeds models with randomly sampled points from the scene, but the sampling process in FS-PCS is non-uniform, favoring more points in the foreground than in the background. This leads to foreground leakage, a noticeable density bias toward foreground classes. This leakage allows previous models to exploit density disparities for easier segmentation, sidestepping the need to learn essential knowledge adaptation patterns for novel classes. Consequently, this issue renders the current benchmark unable to reflect the true performance of previous models.
(2) The second issue is the \textit{sparse point distribution}. 
The current setting samples only 2,048 points during both training and inference due to the huge computational burden in the label propagation module adopted by many FS-PCS methods~\cite{zhao2021few,wang2023few,ning2023boosting}. 
However, this sparse input distribution limits the semantic information available to models, hindering effective advances to improve their recognition ability. 
In addition, this input deviation from real-world scenes diminishes the overall value of research progress in this domain.

To steer the research in the right direction, we standardize the FS-PCS task by proposing a more rigorous setting. Specifically, we correct the foreground leakage and improve the framework by enabling the models to {process a large number of points}, aligning it more closely with real-world scenes. In this well-justified setting, we systematically reevaluate existing methods, establishing a new valid benchmark for future research.

We further introduce a novel FS-PCS model, named \textbf{Correlation Optimization Segmentation (\ourmodel)}. 
As shown in \cref{fig:intro}, existing FS-PCS models are based on \textit{feature optimization}, which means that they optimize support features to enhance prototypes \cite{he2023prototype,ning2023boosting,zhu2023cross,mao2022bidirectional,wang2023few} or optimize query features through fine-grained interaction with support features~\cite{zhang2023few}. 
Instead of operating on features, we propose to optimize the \textbf{Class-specific Multi-prototypical Correlation (CMC)} computed for each query point, representing its correlations to all category prototypes. 
This new \textit{correlation optimization} paradigm allows direct shaping of relationships between query points and category prototypes, leading to better generalization for FS-PCS than feature optimization. 
Building on CMC, we introduce the \textbf{Hyper Correlation Augmentation (HCA)} module. This module refines correlations in the hyperspace by actively interacting them across points and category prototypes.

Moreover, within the meta-learning framework~\cite{vinyals2016matching,snell2017prototypical,ren2018meta} employed by FS-PCS, models undergo training on seen/base classes and are evaluated on unseen/novel classes, revealing an inherent susceptibility. Specifically, these models tend to be susceptible to the familiar base classes within the test scenes, thereby hindering the accurate segmentation of novel classes~\cite{lang2022learning}. To alleviate this susceptibility, we propose a novel approach: learning prototypes for the base classes in a non-parametric and momentum-driven manner during the training phase. Our introduced \textbf{Base Prototypes Calibration (BPC)} module utilizes these learned base prototypes to calibrate correlations for the background within HCA. This calibration effectively mitigates the \textit{base susceptibility} problem, enhancing the model's accuracy.

We systematically benchmark existing methods in our well-justified setting and compare \ourmodel~against others on the S3DIS~\cite{armeni20163d} and ScanNet~\cite{dai2017scannet} datasets (\cref{sec:res}). Our experiments not only reveal the adverse impact of the previous task setting but also highlight the impressive performance of our method.
With extensive ablation studies in \cref{sec:abl}, we offer further insights into the efficacy of our designs and showcase the superior capabilities of the CMC paradigm for FS-PCS, shedding light on future research.

In summary, our contributions include:
\begin{itemize}
\item We identify two significant issues in the current FS-PCS setting: the \textit{foreground leakage} and \textit{sparse point distribution}, which are standardized by our introduction of a rigorous setting and a new benchmark.
\item We propose a novel \textit{correlation optimization} paradigm operating on Class-specific Multi-prototypical Correlation (CMC), enabling the direct shaping of categorical relationships for query points using the Hyper Correlation Augmentation (HCA) module.
\item We tackle the \textit{base susceptibility} issue inherent in meta-learning by introducing non-parametric base prototypes, along with the Base Prototypes Calibration (BPC) module, to calibrate correlations for the background class.
\end{itemize}

\begin{figure*}[!ht]
    \centering
    \includegraphics[width=0.95\linewidth]{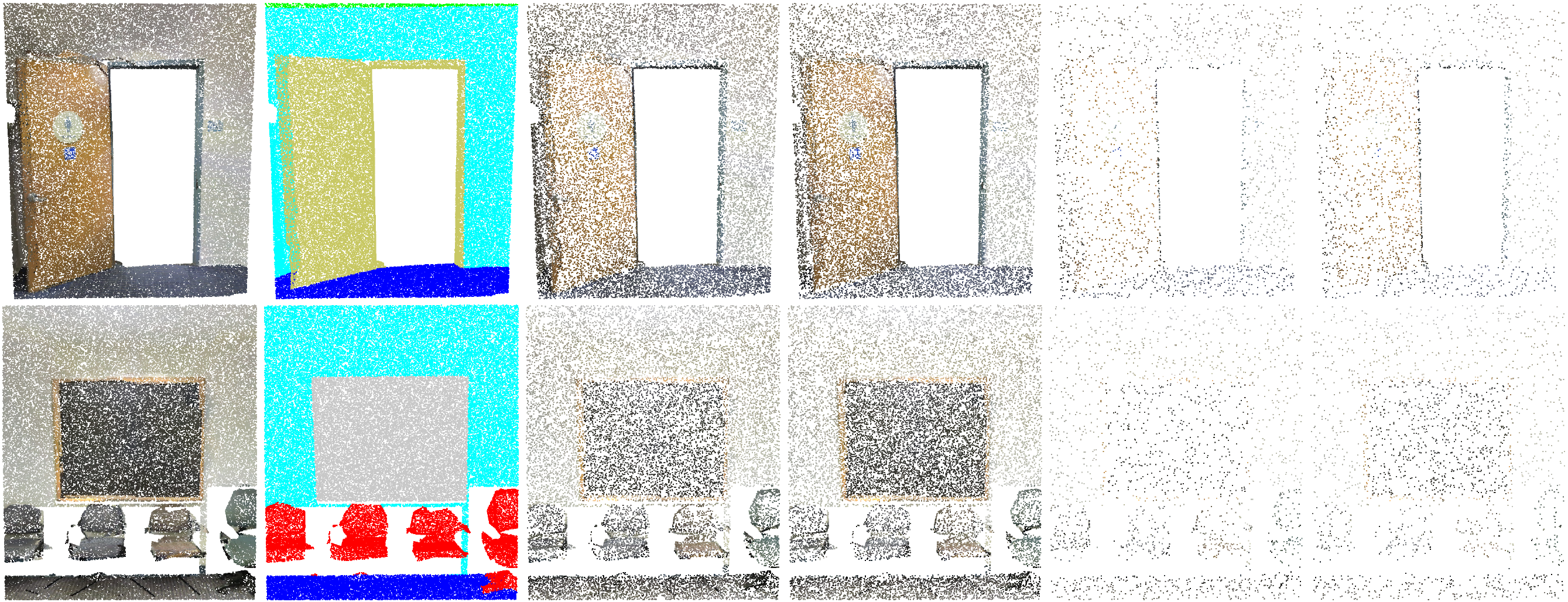}
    \vspace{-0.08in}
    \caption{\textbf{Visualization of two scenes from the S3DIS dataset~\cite{armeni20163d}, with the foreground class as \textit{door} and \textit{board} for 1-way segmentation, respectively.} Each scene includes six types of point clouds, arranged \textit{from left to right:}
    (1) The original point cloud;
    (2) Ground truth of all categories;
    (3) Our corrected input with 20,480 points in a uniform distribution;
    (4) Input with 20,480 points in a biased distribution;
    (5) Input with 2,048 points in a uniform distribution;
    (6) Input with 2,048 points in a biased distribution, as adopted by previous works.}
    \label{fig:sampling}
    \vspace{-0.1in}
\end{figure*}

\section{Related Work}
\subsection{3D Point Cloud Semantic Segmentation}
Currently, numerous approaches have emerged for performing point cloud semantic segmentation in a fully supervised manner, categorized into three main groups. The first group, known as MLP-based methods~\cite{ye20183d,jiang2018pointsift,engelmann2017exploring,qi2017pointnet,qi2017pointnet++,engelmann2018know,hu2020randla}, adopts a shared multi-layer perceptron (MLP) as the core building block, complemented by symmetric functions for aggregating features. In the second group, point convolution-based models~\cite{li2018pointcnn,komarichev2019cnn,thomas2019kpconv,tatarchenko2018tangent,zhang2019shellnet,liu2019point,xu2018spidercnn,atzmon2018point,liu2019relation,hua2018pointwise} adapt convolution kernels to the underlying local geometries due to the unordered nature of point clouds, 
resulting in variations in kernel adaptation techniques. A subset of them~\cite{monti2017geometric,shen2018mining,wang2019dynamic,qi20173d,lei2020spherical,verma2018feastnet,wang2019graph,lin2020convolution,zhou2021adaptive,zhang2023improving} embraces graph-based representations to mirror the structure of point clouds. They employ graph convolutions~\cite{kipf2016semi} to propagate and aggregate features across the graph. The third group incorporates attention mechanisms~\cite{vaswani2017attention} to model long-range dependencies, suitable for handling point cloud irregularities. As a result, various efforts~\cite{zhao2021point,nie2022pyramid,lai2022stratified,ran2021learning,park2022fast,zhang2022patchformer} have been dedicated to leveraging attention mechanisms for feature learning in point cloud segmentation. Notably, Stratified Transformer~\cite{lai2022stratified} proposes a stratified sampling strategy within the self-attention module to enlarge the receptive field without incurring significant computational costs.

\subsection{Few-shot 3D Point Cloud Segmentation}
Given the challenging and labor-intensive nature of point cloud data collection, the importance of FS-PCS becomes increasingly apparent.
The pioneering work, attMPTI~\cite{zhao2021few}, employs label propagation to exploit relationships among prototypes and query points. 
Subsequent works further expand on this foundation. PAP~\cite{he2023prototype} addresses large intra-class feature variations by directly adapting prototypes into the query feature space. QGE~\cite{ning2023boosting} adapts background prototypes to match the query context, followed by the holistic rectification of prototypes under the guidance of query features. 2CBR~\cite{zhu2023cross} leverages co-occurrence features of support and query to calculate bias terms and rectify differences between them. BFG~\cite{mao2022bidirectional} introduces bidirectional feature globalization, activating global perception of prototypes and point features to better aggregate context information. CSSMRA~\cite{wang2023few} develops a multi-resolution attention module using both the nearest and farthest points to enhance context aggregation. SCAT~\cite{zhang2023few} proposes a stratified class-specific attention-based transformer, constructing fine-grained relationships between support and query features. Notably, these methods all pivot on \textit{feature optimization}, refining either support or query features. In contrast, our approach introduces \textit{correlation optimization} by refining the multi-prototypical support-query correlations, which exhibits superior generalization capabilities to novel classes compared to previous feature optimization.

\section{FS-PCS Overview}
\subsection{Task Description}
FS-PCS involves segmenting the foreground semantic categories in a query point cloud as specified by densely annotated support point clouds. 
Formally, following the episodic paradigm~\cite{vinyals2016matching}, each episode for an $N$-way $K$-shot segmentation task contains a support set $\mathcal{S}=\big\{\{\matdn{X}{s}^{n,k},\matdn{Y}{s}^{n,k}\}_{k=1}^{K}\big\}_{n=1}^N$ and a query set $\mathcal{Q}=\{\matdn{X}{q}^n,\matdn{Y}{q}^n\}_{n=1}^N$, where $\matud{X}{s/q}{*}$ and $\matud{Y}{s/q}{*}$ represent a point cloud and its corresponding segmentation mask. 
Within $\mathcal{S}$, each $K$-shot group $\{\matdn{X}{s}^{n,k},\matdn{Y}{s}^{n,k}\}_{k=1}^{K}$ exclusively describes the $n$-th semantic class of the total $N$ foreground classes.

The objective, given $\mathcal{S}$ and $\matdn{X}{q}^n$, is to predict query masks that closely match $\matdn{Y}{q}^n$ by leveraging the knowledge of the $N$ novel categories provided by $N\times K$ support pairs in $\mathcal{S}$. Two semantic category subsets $C_{\rm train}$ and $C_{\rm test}$, with $C_{\rm train} \cap C_{\rm test} = \phi$, are employed for training and testing, respectively.
Thus, training exclusively utilizes foreground classes from $C_{\rm train}$, while testing employs previously unseen classes from $C_{\rm test}$.
The training in FS-PCS usually encompasses two stages, \ie, pre-training and meta-training. The former primarily trains the backbone to learn meaningful semantic features in a fully-supervised manner, while the following meta-training focuses on training the model to transfer knowledge from the support $\mathcal{S}$ to the query $\mathcal{Q}$. Both meta-training and testing adhere to the episodic paradigm.

\begin{figure*}[!ht]
    \centering
    \includegraphics[width=.95\linewidth]{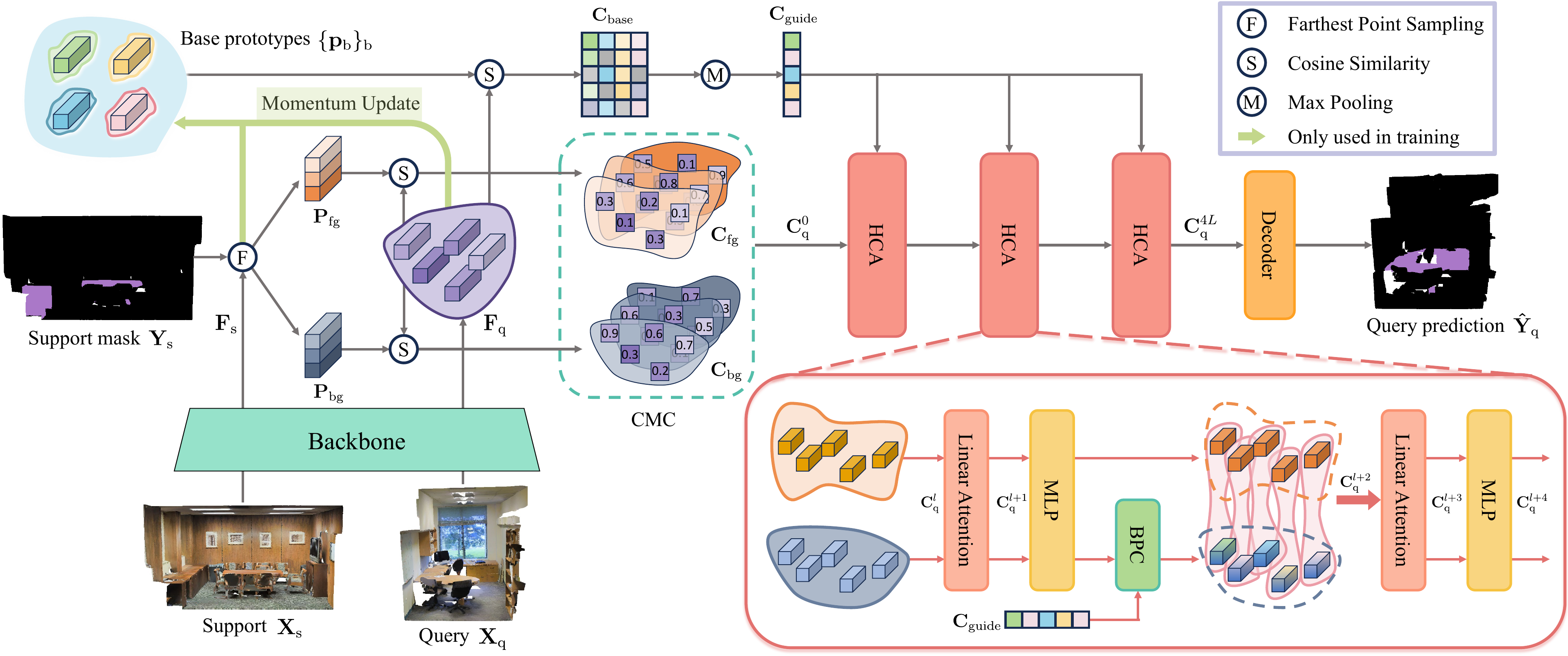}
    \vspace{-0.15in}
    \caption{Overall architecture of the proposed~\ourmodel. Initially, we compute CMC for each query point using the backbone features. 
These correlations are then forwarded to the subsequent HCA module, which actively mines hyper-relations among correlations across points and classes.
Additionally, we dynamically learn non-parametric base prototypes on the fly and introduce the BPC module to effectively alleviate the \textit{base susceptibility} problem. For clarity, we present the model under the 1-way 1-shot setting.
    }
    \label{fig:arch}
    \vspace{-0.20in}
\end{figure*}

\subsection{Issues in the Current Setting} \label{sec:pbs}
The prevailing FS-PCS task setting, initially introduced by~\cite{zhao2021few}, has been consistently employed in subsequent works~\cite{he2023prototype,ning2023boosting,zhu2023cross,mao2022bidirectional,wang2023few,zhang2023few}. Despite the previous progress, we identify two crucial issues within this setting.

\myPara{Foreground Leakage.}
The prevailing 3D segmentation methodology~\cite{lai2022stratified,zhao2021point} feeds models with randomly sampled points from the scene. However, in the current FS-PCS setting, the sampling process introduces a bias toward foreground classes. Specifically, this non-uniform sampling favors foreground classes by sampling more points for them compared to the background, leading to a noticeable point density disparity between foreground and background, thereby leaking the foreground classes to the models. More details on this biased sampling are available in the supplementary material. As depicted in~\cref{fig:sampling}, the inputs (3), (5) from the corrected uniform sampling show balanced point distributions, while the inputs (4), (6) using biased sampling exhibit denser distributions in the foreground (\textit{door} or \textit{board}) than in the background. This foreground leakage induces models to segment foreground classes by identifying denser regions, instead of learning semantic knowledge transfer from support to query. This issue, occurring in both training and testing, undermines the benchmark's validity. Addressing this issue, as shown in~\cref{sec:res}, unveils a significant performance drop in existing methods, emphasizing the imperative need for correction.

\myPara{Sparse Point Distribution.}
Besides, the current FS-PCS input is constrained to only 2,048 points due to the high computational cost of constructing a $k$-nearest neighbor graph in the label propagation module adopted by many FS-PCS methods~\cite{zhao2021few,wang2023few,ning2023boosting}.
However, this sparse point distribution severely limits semantic clarity, making it difficult to distinguish objects. 
For instance, in~\cref{fig:sampling}, it is even challenging for humans to distinguish the \textit{door} and surrounding \textit{wall} in the 2048-point input ($5^{th}$ column) in the $1^{st}$ row. 
The same applies to the $2^{nd}$ row to discern \textit{board} from other classes like \textit{window}. 
These sparsely populated, semantically limited inputs introduce significant ambiguities, hindering the model's capacity to exploit semantics in the scenes. 
Furthermore, this deviation from real-world scenes limits the scope of current research progress.

To address these issues, we introduce a more rigorous setting for FS-PCS. In this standardized setting, we increase the number of input points tenfold to 20,480 and eliminate foreground leakage through uniform sampling. 
As depicted in~\cref{fig:sampling}, the input (3) from this rigorous setting provides clearer scene representations and uniform distributions, aligning the task setting more closely with real-world scenarios. 
The new benchmark results under this setting are presented in~\cref{sec:res}.

\section{Methodology}
\label{sec:meth}
Instead of employing the traditional \textit{feature optimization} \cite{he2023prototype,ning2023boosting,zhu2023cross,mao2022bidirectional,wang2023few,zhang2023few}, our proposed \ourmodel~is built upon the \textit{correlation optimization} paradigm with CMC, allowing for direct refinement of relationships between each query point and category prototypes. 
\cref{fig:arch} illustrates the pipeline of~\ourmodel.
Without loss of generality, we present our model under the 1-way 1-shot setting in the following sections.

\subsection{Class-specific Multi-prototypical Correlation}
Given the backbone $\Phi$, we extract support features $\matdn{F}{s} = \Phi(\matdn{X}{s}) \in \mathbb{R}^{N_S \times D}$ and query features $\matdn{F}{q} = \Phi(\matdn{X}{q}) \in \mathbb{R}^{N_Q \times D}$, where $D$ is the channel dimension, $N_S$ and $N_Q$ are the number of points in $\matdn{X}{s}$ and $\matdn{X}{q}$, respectively. 
Foreground prototypes $\matdn{P}{fg}$ and background prototypes $\matdn{P}{bg}$ are obtained through two steps: sample $N_O$ seeds in the coordinate space based on farthest point sampling, and then conduct point-to-seed clustering~\cite{zhao2021few} as follows:
\begin{align}
\small
\begin{split}
\matdn{P}{fg} &= \mathcal{F}_{{\rm clus}}(\matdn{F}{s} \odot \matdn{Y}{s}, \matdn{S}{fg}),\ \matdn{S}{fg} = \mathcal{F}_{{\rm fps}}(\matdn{L}{s}\odot \matdn{Y}{s}),\\
\matdn{P}{bg} &= \mathcal{F}_{{\rm clus}}(\matdn{F}{s} \odot \matdn{\Tilde{Y}}{s}, \matdn{S}{bg}),\ \matdn{S}{bg} = \mathcal{F}_{{\rm fps}}(\matdn{L}{s}\odot \matdn{\Tilde{Y}}{s}),
\end{split}
\end{align}
where $\odot$ is the Hadamard product, $\matdn{L}{s}$ denotes the $xyz$ coordinates of support points from $\matdn{X}{s}$, $\matdn{\Tilde{Y}}{s}$ is the inverse mask of $\matdn{{Y}}{s}$, and $\mathcal{F}_{{\rm fps}}$ represents the farthest point sampling operation.
Continuously, $\matdn{S}{fg/bg}$ is the set of indices corresponding to the seeds sampled by $\mathcal{F}_{{\rm fps}}$, and $\mathcal{F}_{{\rm clus}}$ stands for the clustering operation. After this, we have $\matdn{P}{fg}, \matdn{P}{bg} \in \mathbb{R}^{N_O\times D}$ with $N_O$ prototypes per category.

Next, we compute the cosine similarities of query points with respect to $\matdn{P}{fg}$ and $\matdn{P}{bg}$, and obtain the correlations $\matdn{C}{fg} \in \mathbb{R}^{N_Q\times N_O}$ and $\matdn{C}{bg} \in \mathbb{R}^{N_Q\times N_O}$, given by:
\begin{align}
\small
\label{eq:cs}
\matdn{C}{fg} = \dfrac{\matdn{F}{q} \cdot \matdn{P}{fg}^\intercal}{\left\| \matdn{F}{q}\right\| \left\| \matdn{P}{fg}^\intercal\right\| },\ 
\matdn{C}{bg} = \dfrac{\matdn{F}{q} \cdot \matdn{P}{bg}^\intercal}{\left\| \matdn{F}{q}\right\| \left\| \matdn{P}{bg}^\intercal\right\| }.
\end{align}

Finally, the correlations $\matdn{C}{fg}$ and $\matdn{C}{bg}$ are both expanded to the size of $\mathbb{R}^{N_Q\times 1\times N_O}$. We concatenate them along the second dimension and project the last dimension back to $D$ using an MLP $\mathcal{F}_{{\rm mlp}}$, as follows:
\begin{align}
\small
\label{eq:proj}
\matdn{C}{q}^0 = \mathcal{F}_{{\rm mlp}}(\matdn{C}{fg}\oplus \matdn{C}{bg}) \in \mathbb{R}^{N_Q\times N_C\times D}, 
\end{align}
where $\oplus$ is the concatenation operation. \cref{eq:proj} yields the initial CMC.
Notably, the second dimension $N_C$ of $\matdn{C}{q}^0$ is the number of classes, which is $2$ under this $1$-way example. 

The initial correlations $\matdn{C}{q}^0$ comprise the correlations of each query point with a number of prototypes for all classes, which allows the subsequent modules to directly shape the relations between the query and support. Such \textit{correlation optimization} leads to enhanced generalization for FS-PCS compared to the traditional \textit{feature optimization} \cite{he2023prototype,ning2023boosting,zhu2023cross,mao2022bidirectional,wang2023few,zhang2023few}, as demonstrated in~\cref{sec:abl}.

\subsection{Hyper Correlation Augmentation}
Our proposed CMC denotes the correlations of each query point to all category prototypes. 
To enhance the correlations, we introduce the Hyper Correlation Augmentation (HCA) module, leveraging two underlying relationships. 
First, the query points are all related and dependent on each other. Their correlations to all prototypes are also connected, leading to point-point relations. 
Second, classifying a single point into foreground or background depends on its \textit{relative} correlations to foreground or background prototypes, forming foreground-background relations. 
For an $N$-way setting, this extends to foregrounds-background relations, considering the \textit{relative} correlations among all classes.
The proposed HCA refines correlations by exploiting both point-point and foreground-background relations.

\noindent\textbf{Linear Attention.} 
Due to the irregular nature of 3D point clouds, the attention mechanism with the permutation-invariant property is well-suited for point cloud processing. Here, we adopt linear attention~\cite{katharopoulos2020transformers} for its global receptive field and superior linear computation efficiency.

Given an input sequence $\mathbf{C} \in \mathbb{R}^{N\times D}$, applying linear transformations to $\mathbf{C}$ results in $\mathbf{Q}, \mathbf{K}, \mathbf{V} \in \mathbb{R}^{N\times D}$. Using $\mathbf{q}_i$, $\mathbf{k}_i$, and $\mathbf{v}_i \in \mathbb{R}^{1\times D}$ to denote the $i$-th token vector from $\mathbf{Q}$, $\mathbf{K}$, and $\mathbf{V}$, respectively, the standard attention~\cite{vaswani2017attention} is:
\begin{align}
\small
\label{eq:stdattn}
\mathbf{\hat{v}}_i = \frac{\sum_{j=1}^N\langle \mathbf{q}_i, \mathbf{k}_j\rangle \mathbf{v}_j}{\sum_{j=1}^N\langle \mathbf{q}_i, \mathbf{k}_j\rangle}\ {,}\ \langle \mathbf{q}_i, \mathbf{k}_j\rangle = {\rm exp}(\frac{\mathbf{q}_i\mathbf{k}_j^\intercal}{\sqrt{D}}),
\end{align}
where $\langle \cdot, \cdot \rangle$ represents the similarity measure function.

Through the lens of kernels, linear attention defines $\langle \mathbf{q}_i, \mathbf{k}_j\rangle = \varphi(\mathbf{q}_i)\varphi(\mathbf{k}_j)^\intercal$ in \cref{eq:stdattn} and utilizes the associative property of matrix multiplication to obtain:
\begin{align}
\small
\label{eq:linattn}
\mathbf{\hat{v}}_i = \frac{\varphi(\mathbf{q}_i)\sum_{j=1}^N\varphi(\mathbf{k}_j)^\intercal\mathbf{v}_j}{\varphi(\mathbf{q}_i)\sum_{j=1}^N\varphi(\mathbf{k}_j)^\intercal},
\end{align}
where $\varphi(x) = {\rm elu}(x) + 1$ with ${\rm elu}(\cdot)$ as the exponential linear unit~\cite{clevert2015fast}.
Consequently, the computation cost of linear attention is $\mathcal{O}(ND^2)$, significantly more favorable than the standard $\mathcal{O}(N^2D)$ complexity.

\noindent\textbf{Hyper Correlation Augmentation.} {Based on linear attention, we introduce the HCA module to enhance the correlations through active interactions across points and classes.}
Since we stack the HCA module $L$ times as in~\cref{fig:arch}, the module input is denoted as $\matdn{C}{q}^l$.
For each point, we first attend its correlations with those of all other points. We permute $\matdn{C}{q}^l$ with the class dimension as its first dimension and then compute linear attention across points:
\begin{align}
\small
\label{eq:pib1}
\matdn{C}{q}^{l+1} = \mathcal{F}_{\rm lnatt}(\mathcal{T}(\matdn{C}{q}^l)) \in \mathbb{R}^{N_C\times N_Q\times D},
\end{align}
where $\mathcal{T}$ transposes the first two dimensions, and $\mathcal{F}_{\rm lnatt}$ represents the linear attention layer to process features independently along the first dimension.
Following the attention layer, an MLP is applied to each point separately and identically to further enhance the correlations:
\begin{align}
\small
\label{eq:pib2}
\matdn{C}{q}^{l+2} = \mathcal{F}_{{\rm mlp}} (\matdn{C}{q}^{l+1})  \in \mathbb{R}^{N_C\times N_Q\times D}.
\end{align}
Note that multi-head attention~\cite{vaswani2017attention}, layer normalization~\cite{ba2016layer}, and residual connections are omitted here for simplicity.

After that, we leverage the foreground-background relations to facilitate learning categorical relationships and determining the best-fit class for each point's semantics.
We rearrange the dimensions such that $\matdn{C}{q}^{l+2} \in \mathbb{R}^{N_C\times N_Q\times D} \rightarrow \mathbb{R}^{N_Q\times N_C\times D}$, and apply linear attention, given by:
\begin{align}
\small
\label{eq:cib}
\matdn{C}{q}^{l+3} = \mathcal{F}_{\rm lnatt}(\mathcal{T}(\matdn{C}{q}^{l+2})) \in \mathbb{R}^{N_Q\times N_C\times D}.
\end{align}
The next MLP transforms $\matdn{C}{q}^{l+3}$ to $\matdn{C}{q}^{l+4}$ as in~\cref{eq:pib2}.

Through this module, CMC can interact not only across the spatial dimension but also across the categorical space. This results in comprehensive contextual dependencies, significantly enhancing meta-learning performance.

\subsection{Base Prototypes Calibration}
Since the training concentrates on classes in $C_{\rm train}$, models are inherently biased towards these base classes, hindering the segmentation of novel classes~\cite{lang2022learning}. To address it, we propose employing non-parametric prototypes for base classes through the BPC module to alleviate the base bias.

Let $\{\matdn{p}{b}|\matdn{p}{b} \in \mathbb{R}^{1\times D}\}_{{\rm b}=1}^{N_b}$ be a set of non-parametric prototypes corresponding to $N_b = |C_{\rm train}|$ base classes. During meta-training, these prototypes are zero-initialized and evolve continuously.
Specifically, given $\matdn{F}{s}$, $\matdn{F}{q}$, and their binary annotations $\matud{Y}{s}{b}$, $\matud{Y}{q}{b}$ of the $b$-th base class, we calculate the Masked Average Pooling (MAP)~\cite{zhang2020sg} for each base class present in the current point clouds:
\begin{align}
\small
\label{eq:bpc1}
\matdn{p}{b}' = \mathcal{F}_{\rm pool}(\matdn{F}{s/q}  {\odot} \matud{Y}{s/q}{b}) \in \mathbb{R}^{1\times D},
\end{align}
where $\mathcal{F}_{\rm pool}$ represents the MAP operation.
Then, the base prototypes can be updated at each training episode as:
\begin{align}
\small
\label{eq:bpc2}
\matdn{p}{b} \leftarrow \mu\matdn{p}{b} + (1 - \mu)\matdn{p}{b}',
\end{align}
where $\mu \in [0,1]$ is a momentum coefficient.

When segmenting the novel classes, the query point corresponding to the base classes should be considered as background. Therefore, leveraging base prototypes, we introduce the BPC module to calibrate correlations to the background class, mitigating potential interference from base susceptibility. 
Specifically, we calculate the base correlations $\matdn{C}{base}$ between the query and base prototypes:
\begin{align}
\small
\label{eq:bpc3}
\matdn{C}{base} = \dfrac{\matdn{F}{q} \cdot {\mathcal{I}(\{\matdn{p}{b}\}_{\rm b=1}^{N_b})}^\intercal}{\left\| \matdn{F}{q}\right\| \left\| {\mathcal{I}(\{\matdn{p}{b}\}_{\rm b=1}^{N_b})}^\intercal\right\|} \in \mathbb{R}^{N_Q \times N_b},
\end{align}
where $\mathcal{I}$ concatenates all the vectors in the set, such that $\mathcal{I}(\{\matdn{p}{b}\}_{\rm b=1}^{N_b}) \in \mathbb{R}^{N_b \times D}$. 
Afterward, we obtain the base guidance for each query point $\matdn{C}{guide} = \mathcal{F}_{\rm max}(\matdn{C}{base}) \in \mathbb{R}^{N_Q}$, where $\mathcal{F}_{\rm max}$ is max pooling on each row in $\matdn{C}{base}$. 
Then, the background correlations are calibrated by $\matdn{C}{guide}$ before interacting with foreground correlations, 
as in~\cref{fig:arch}:
\begin{align}
\small
\label{eq:bpc4}
\matdn{C}{q}^{l+2}[1,\cdot,\cdot] = \mathcal{F}_{{\rm fc}} (\matdn{C}{q}^{l+2}[1,\cdot,\cdot]\oplus \mathcal{D}(\matdn{C}{guide})),
\end{align}
where $\matdn{C}{q}^{l+2}[1,\cdot,\cdot] \in \mathbb{R}^{N_Q\times D}$ selects the background correlations (the last at the $N_C$-dim) from CMC, $\mathcal{D}$ expands $\matdn{C}{guide}$ as $\mathbb{R}^{N_Q} \rightarrow \mathbb{R}^{N_Q\times D}$, and $\mathcal{F}_{{\rm fc}}$ is a fully connected layer. 
During meta-training, we exclude the base prototypes of the current target classes in~\cref{eq:bpc3}. 
For evaluation, the base prototypes are frozen and utilized without exclusion.

Finally, $\matdn{C}{q}^{4L}$ is decoded to the final segmentation result $\matdn{\hat{Y}}{q}$ using the decoder. Another MLP is employed to generate base class predictions using query features $\matdn{F}{q}$. The entire model is optimized using cross-entropy (CE) loss:
\begin{align}
\small
\label{eq:bpc5}
\mathcal{L} = {\rm CE}(\mathcal{F}_{{\rm mlp}}(\matdn{F}{q}),\{\matud{Y}{q}{b}\}_{\rm b}) + {\rm CE}(\matdn{\hat{Y}}{q},\matdn{Y}{q}).
\end{align}

\begin{table*}[!t]
\centering
\setlength{\tabcolsep}{2.0mm}
\renewcommand{\arraystretch}{.9}
\resizebox{\linewidth}{!}{
\begin{tabular}{c|l|ccccccccccccccc} \toprule
    & \multirow{2}{*}{Methods} & \multicolumn{3}{c}{$1$-shot (S3DIS)} & \phantom{a} & \multicolumn{3}{c}{$5$-shot (S3DIS)} & \phantom{a} & \multicolumn{3}{c}{$1$-shot (ScanNet)} & \phantom{a} & \multicolumn{3}{c}{$5$-shot (ScanNet)}\\ 
    \cmidrule{3-5} \cmidrule{7-9} \cmidrule{11-13} \cmidrule{15-17}
    & & $S^0$  & $S^1$  & mean & \phantom{a} & $S^0$  & $S^1$  & mean & \phantom{a} & $S^0$  & $S^1$  & mean & \phantom{a} & $S^0$  & $S^1$  & mean  \\ \midrule 
    \multirow{3}{*}{\textit{w/ FG}} 
    & AttMPTI~\cite{zhao2021few}        & 64.89 & 66.15 & 65.52 && 76.56 & 83.08 & 79.82 && 62.14 & 58.65 & 60.39 && 68.79 & 68.66 & 68.73 \\
    & QGE~\cite{ning2023boosting}       & 74.05 & 73.61 & 73.83 && 74.65 & 83.21 & 78.93 && 63.50 & 57.61 & 60.56 && 70.72 & 65.68 & 68.20 \\
    & QGPA~\cite{he2023prototype}       & 62.72 & 61.95 & 62.33 && 76.30 & 87.29 & 81.80 && 56.47 & 51.72 & 54.10 && 81.57 & 72.75 & 77.16 \\ 
    \midrule
    \multirow{3}{*}{\textit{w/o FG}}
    & AttMPTI~\cite{zhao2021few}        & 41.56 & 41.27 & 41.41 && 50.55 & 46.13 & 48.34 && 33.36 & 31.81 & 32.58 && 37.95 & 36.30 & 37.12 \\
    & QGE~\cite{ning2023boosting}       & 46.27 & 47.76 & 47.02 && 47.74 & 59.77 & 53.76 && 37.72 & 34.64 & 36.18 && 48.73 & 39.95 & 44.34 \\
    & QGPA~\cite{he2023prototype}       & 35.62 & 41.13 & 38.38 && 43.54 & 47.50 & 45.52 && 40.03 & 35.54 & 37.78 && 46.17 & 42.24 & 44.20\\ 
    \bottomrule
\end{tabular}}
\vspace{-0.10in}
\caption{\textbf{Comparisons in the mIoU metric between \textit{with} foreground leakage (\textit{w/ FG}) and \textit{without} foreground leakage (\textit{w/o FG}) for existing methods.} The results are for $1$-way segmentation setting. $S^0$/$S^1$ refers to the $i$-th split for inference. Here, we adopt the previous FS-PCS setting \cite{zhao2021few}, \ie, using DGCNN~\cite{wang2019dynamic} as the backbone and sampling 2,048 points for each scene.}
\label{table:setting}
\vspace{-0.10in}
\end{table*}

\begin{table*}[!t]
\centering
\setlength{\tabcolsep}{2.05mm}
\renewcommand{\arraystretch}{.9}
\newcommand{\gray}{\cellcolor{mygray}}
\resizebox{\linewidth}{!}{
\begin{tabular}{c|l|ccccccccccccccc} \toprule
    & \multirow{2}{*}{Methods} & \multicolumn{3}{c}{$1$-way $1$-shot} & \phantom{a} & \multicolumn{3}{c}{$1$-way $5$-shot} & \phantom{a} & \multicolumn{3}{c}{$2$-way $1$-shot} & \phantom{a} & \multicolumn{3}{c}{$2$-way $5$-shot}\\ 
    \cmidrule{3-5} \cmidrule{7-9} \cmidrule{11-13} \cmidrule{15-17}
    & & $S^0$  & $S^1$  & mean & \phantom{a} & $S^0$  & $S^1$  & mean & \phantom{a} & $S^0$  & $S^1$  & mean & \phantom{a} & $S^0$  & $S^1$  & mean  \\ \midrule 
    \multirow{4}{*}{S3DIS~\cite{armeni20163d}}
    & AttMPTI~\cite{zhao2021few}        & 36.32 & 38.36 & 37.34 && 46.71 & 42.70 & 44.71 && 31.09 & 29.62 & 30.36 && 39.53 & 32.62 & 36.08 \\
    & QGE~\cite{ning2023boosting}       & 41.69 & 39.09 & 40.39 && 50.59 & 46.41 & 48.50 && 33.45 & 30.95 & 32.20 && 40.53 & 36.13 & 38.33 \\
    & QGPA~\cite{he2023prototype}       & 35.50 & 35.83 & 35.67 && 38.07 & 39.70 & 38.89 && 25.52 & 26.26 & 25.89 && 30.22 & 32.41 & 31.32 \\ 
    & \gray\ourmodel~(ours) & \gray\textbf{46.31} & \gray\textbf{48.10} & \gray\textbf{47.21} & \gray  & \gray\textbf{51.40} & \gray\textbf{48.68} & \gray\textbf{50.04} & \gray & \gray\textbf{37.44} & \gray\textbf{36.45} & \gray\textbf{36.95} & \gray & \gray\textbf{42.27} & \gray\textbf{38.45} & \gray\textbf{40.36} \\ \midrule
    \multirow{4}{*}{ScanNet~\cite{dai2017scannet}}
    & AttMPTI~\cite{zhao2021few}        & 34.03 & 30.97 & 32.50 && 39.09 & 37.15 & 38.12 && 25.99 & 23.88 & 24.94 && 30.41 & 27.35 & 28.88 \\
    & QGE~\cite{ning2023boosting}       & 37.38 & 33.02 & 35.20 && 45.08 & 41.89 & 43.49 && 26.85 & 25.17 & 26.01 && 28.35 & 31.49 & 29.92 \\
    & QGPA~\cite{he2023prototype}       & 34.57 & 33.37 & 33.97 && 41.22 & 38.65 & 39.94 && 21.86 & 21.47 & 21.67 && 30.67 & 27.69 & 29.18 \\ 
    & \gray\ourmodel~(ours) & \gray\textbf{41.73} & \gray\textbf{41.82} & \gray\textbf{41.78} & \gray & \gray\textbf{48.31} & \gray\textbf{44.11} & \gray\textbf{46.21} & \gray & \gray\textbf{28.72} & \gray\textbf{28.83} & \gray\textbf{28.78} & \gray & \gray\textbf{35.97} & \gray\textbf{33.39} & \gray\textbf{34.68} \\ \bottomrule
\end{tabular}}
\vspace{-0.10in}
\caption{\textbf{Comparisons in the mIoU metric between our method and baselines in the new FS-PCS setting.} The best-performing results are highlighted in \textbf{bold}. Previous methods apply the same backbone as ours for fair comparisons.}
\label{table:results}
\vspace{-0.10in}
\end{table*}

\begin{figure*}[t!]
    \centering
    \includegraphics[width=.95\linewidth]{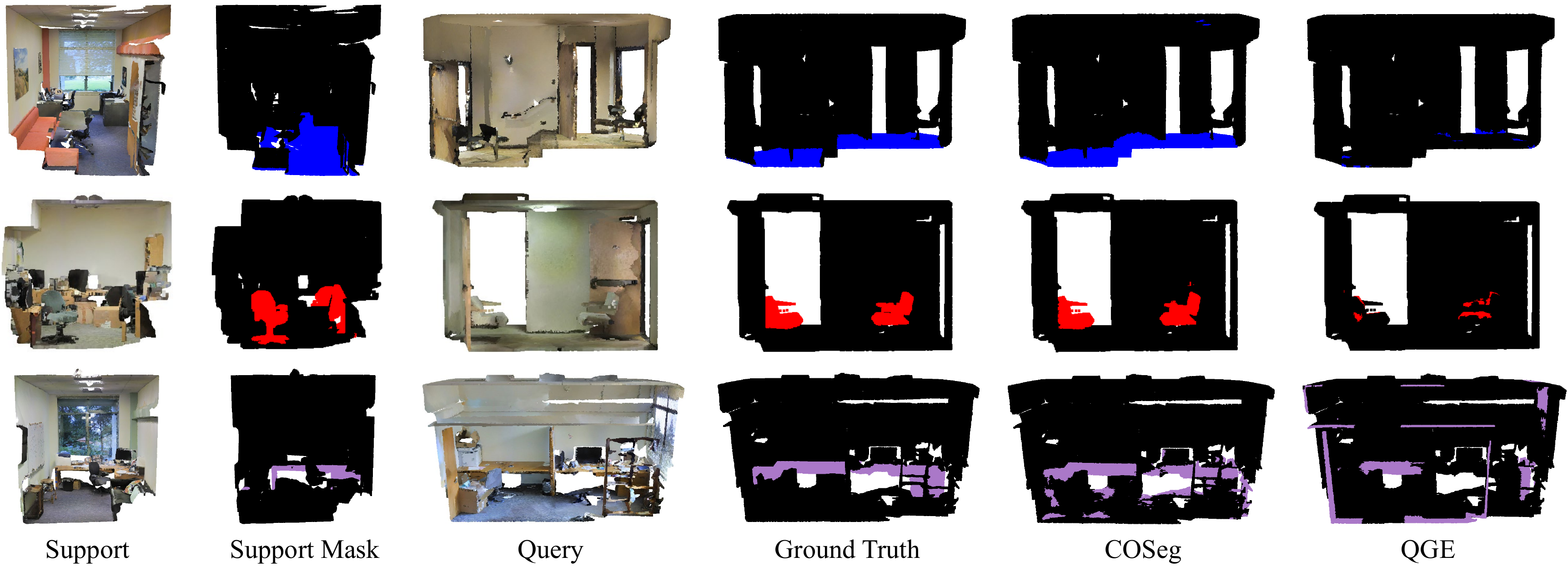}
    \vspace{-0.10in}
    \caption{
    Qualitative comparisons between our proposed model~\ourmodel~and QGE~\cite{ning2023boosting}. 
    Each row, from top to bottom, represents the $1$-way $1$-shot task with the target category as floor (\textcolor{floor}{blue}), chair (\textcolor{chair}{red}), and table (\textcolor{mypurple}{purple}), respectively.
    }
    \label{fig:vs1}
    \vspace{-0.15in}
\end{figure*}

\section{Experiments}
\label{sec:exp}
\subsection{Experimental Setting}
\textbf{Network Architecture.} 
We use the first three blocks from the Stratified Transformer~\cite{lai2022stratified} as our backbone. 
The last two blocks produce features with resolutions 1/4 and 1/16 of the original point cloud. 
We perform interpolation~\cite{qi2017pointnet++} to 4× upsample the 1/16 feature map and concat it to the 1/4 features, followed by an MLP to obtain final features with a channel dimension of 192.
For the S3DIS dataset, we employ a 2-layer HCA module. Due to the richer semantics of ScanNet~\cite{dai2017scannet}, we use 4 layers of HCA. The final decoder consists of one KPConv~\cite{thomas2019kpconv} layer followed by an MLP.

\noindent\textbf{Implementation Detail.} 
We employ the data processing strategy from~\cite{zhao2021few, lai2022stratified}. The 3D scene is divided into 1m $\times$ 1m blocks to increase data samples, and raw input points are grid-sampled with a grid size of 0.02m. 
After voxelization, if the input point count exceeds 20,480, we randomly sample 20,480 points to control the input size. 
Data augmentation and pre-training follow~\cite{lai2022stratified} where our backbone is pre-trained on each fold for 100 epochs. 
Meta-training involves 40,000 episodes, using the AdamW optimizer with a learning rate of 0.00005 and weight decay of 0.01. 
During testing, we sample 1,000 episodes per class in the $1$-way setting and 100 episodes for each combination in the $2$-way setting for more stable evaluations. 
We use 100 prototypes for each class ($N_O=100$). 
In the $k$-shot setting, when $k > 1$, we sample $N_O/k$ prototypes from each shot and concatenate them to obtain $N_O$ prototypes. 
For benchmarking previous models, we select methods with publicly available code, namely AttMPTI~\cite{zhao2021few}, QGE~\cite{ning2023boosting}, and QGPA~\cite{he2023prototype}.

\subsection{Main Results}
\label{sec:res}
\noindent\textbf{Effects of Foreground Leakage.} 
As discussed in~\cref{sec:pbs}, the foreground leakage problem can significantly distort model training and evaluation.~\cref{table:setting} compares the performance of previous methods~\cite{zhao2021few,ning2023boosting,he2023prototype} in two settings: one with foreground leakage (\textit{w/ FG}) and the other without foreground leakage (\textit{w/o FG}) across two datasets.
In each setting, we retrain the models and evaluate them on the corresponding test set. 
The results reveal a substantial mIoU drop after correcting foreground leakage consistently across all splits for $1$-way $1/5$-shot tasks.
On S3DIS, the highest mIoU of 81.80\% (\textit{w/ FG}) for the $5$-shot task drops dramatically to 45.52\% after removing foreground leakage, marking a significant 36.28\% drop. 
Similarly, on ScanNet, a substantial 32.96\% mIoU drop is observed from 77.16\% to 44.20\% for the $5$-shot task.
On average, across the three methods, the mIoU drop from removing foreground leakage is 27.97\% on S3DIS and 26.16\% on ScanNet. 
This notable performance gap underscores that previous methods largely rely on the density differences exposed by foreground leakage to achieve seemingly superior performance. 
This underscores the immediate need for our new corrected setting for facilitating the research in FS-PCS.

\begin{table*}[!t]
    \renewcommand{\arraystretch}{.9}
    \begin{minipage}[c]{.34\textwidth}
        \centering
        \resizebox{\linewidth}{!}{
        \begin{tabular}{lccccc} 
        \toprule
         Optimization   &  HCA  &  BPC & \multicolumn{1}{c}{1-shot} &  \multicolumn{1}{c}{5-shot}\\ 
                                          \midrule 
        feature      & &      & 30.67 & 32.58  \\
        correlation  & &      & 39.93 & 42.33  \\ 
        correlation  & \checkmark &      & 43.77 & 47.98  \\ 
        correlation  & \checkmark & \checkmark & 47.21 & 50.04  \\ 
        \bottomrule
        \end{tabular}}
        \vspace{-0.10in}
        \caption{Ablation study of different design choices in COSeg.}
        \label{table:design}
    \end{minipage}\hfill
    \begin{minipage}[c]{.21\linewidth}
        \centering
        \resizebox{.805\linewidth}{!}{
        \begin{tabular}{lccc} 
        \toprule
         $N_O$          & \multicolumn{1}{c}{1-shot} &  \multicolumn{1}{c}{5-shot}\\ 
                                          \midrule 
        50        & 43.50 & 46.43  \\
        100       & 47.21 & 50.04   \\
        150       & 47.41 & 52.33  \\ 
        200       & 48.27 & 47.90  \\ 
        \bottomrule
        \end{tabular}}
        \vspace{-0.10in}
        \caption{Effects of the number of prototypes per class.}
        \label{table:proto}
    \end{minipage}\hfill
    \begin{minipage}[c]{.18\linewidth}
        \renewcommand{\arraystretch}{1.16}
        \centering
        \resizebox{.84\linewidth}{!}{
        \begin{tabular}{lccc} 
        \toprule
         $L$          & \multicolumn{1}{c}{1-shot} &  \multicolumn{1}{c}{5-shot}\\ 
                                          \midrule 
        1       & 44.19 & 45.93  \\
        2       & 47.21 & 50.04   \\
        3       & 46.12 & 46.89  \\ 
        \bottomrule
        \end{tabular}}
        \vspace{-0.10in}
        \caption{Impact of the number of HCA layers.}
        \label{table:iici}
    \end{minipage}\hfill
    \begin{minipage}[c]{.19\linewidth}
        \renewcommand{\arraystretch}{1.16}
        \centering
        \resizebox{.93\linewidth}{!}{
        \begin{tabular}{lccc} 
        \toprule
         $\mu$          & \multicolumn{1}{c}{1-shot} &  \multicolumn{1}{c}{5-shot}\\ 
                                          \midrule 
        0.99        & 46.91 & 50.26  \\
        0.995       & 47.21 & 50.04   \\
        0.999       & 47.40 & 49.85  \\ 
        \bottomrule
        \end{tabular}}
        \vspace{-0.10in}
        \caption{Ablation on the momentum coefficient.}
        \label{table:mom}
    \end{minipage}
    \vspace{-0.15in}
\end{table*}

\begin{figure*}[!t]
    \centering
    \includegraphics[width=.98\linewidth]{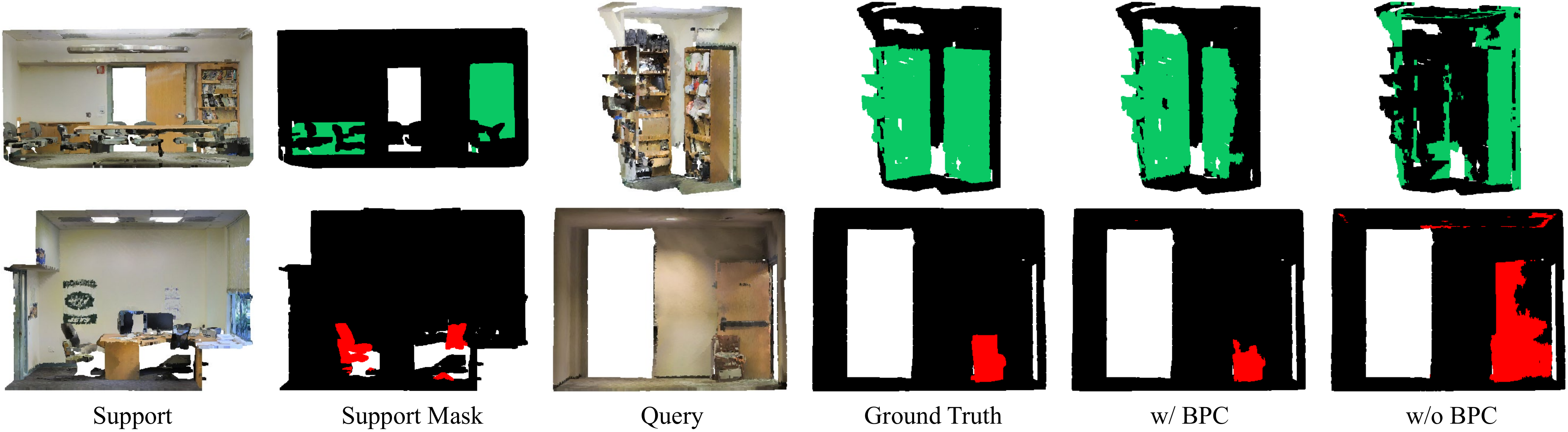}
    \vspace{-0.10in}
    \caption{Visual comparisons between our models with BPC (w/ BPC) and without BPC (w/o BPC). 
    Each row corresponds to the $1$-way $1$-shot task targeting bookcase (\textcolor{bookcase}{green}) and chair (\textcolor{chair}{red}), respectively, arranged from top to bottom.}
    \label{fig:vs2}
    \vspace{-0.15in}
\end{figure*}

\noindent\textbf{Comparison with Previous Methods.}
In~\cref{table:results}, we present the results for $1/2$-way $1/5$-shot experiments on two datasets. 
Our model demonstrates a significant performance advantage, establishing new state-of-the-art records across all experiments.
For instance, in the $1$-way $1$-shot scenario, we achieve notable mIoU improvements of 6.82\% and 6.58\% over the second-best model, QGE, on S3DIS and ScanNet, respectively. 
Extending to the $2$-way $5$-shot task, our model outperforms the previous best performance by 2.03\% (S3DIS) and 4.76\% (ScanNet) in mIoU. 
Similar substantial improvements are observed in all other settings. 
These consistent enhancements underscore the efficacy of our model within our proposed rigorous FS-PCS setting.

\noindent\textbf{Qualitative Results.} 
In~\cref{fig:vs1}, we visualize predictions from our method ($5$th column) and the previous best method, QGE ($6$th column). 
Our method clearly achieves better segmentation results than the previous best method.

\subsection{Ablation Study}
\label{sec:abl}
In this section, we report the mIoU results as \textbf{the mean of all splits} of S3DIS under the $1$-way $1/5$-shot settings.

\noindent\textbf{Different Design Choices.}
We first assess our proposed correlation optimization paradigm. 
The baseline model consists of only the backbone and decoder from our approach. 
We compare the performance of forwarding correlations (correlation optimization) and forwarding features (feature optimization) to the decoder after the backbone. 
Feature optimization uses the support prototype to directly segment the target object as in~\cite{tian2020prior}. 
As shown in~\cref{table:design}, transitioning solely from forwarding features to forwarding correlations results in a significant 9.26\% and 9.75\% increase in mIoU under $1/5$-shot settings, respectively. 
These results affirm the superiority of our proposed correlation optimization paradigm in enhancing generalization for FS-PCS compared to the traditional feature optimization. 

Furthermore, we explore the impact of HCA and BPC in~\cref{table:design}. 
Adding HCA to the baseline with correlation optimization leads to a 3.84\%/5.65\% mIoU improvement for the $1$-shot/$5$-shot setting, demonstrating the efficacy of HCA in enriching contextual information for correlations. 
Incorporating BPC with HCA results in an additional 3.44\%/2.06\% growth in mIoU for the $1$-shot/$5$-shot setting, highlighting the significance of BPC in calibrating background correlations.
\cref{fig:vs2} contrasts visual segmentation results between our models with BPC and without BPC.
The absence of BPC exhibits base susceptibility issues, with false activations of base classes (\textit{wall} or \textit{door}) in the scenes. 
Conversely, the inclusion of our BPC design enables models to effectively mitigate susceptibility, ensuring accurate segmentation of novel classes.

\noindent\textbf{Number of Prototypes.}
\cref{table:proto} shows increasing the number of prototypes to $150$ improves performance. For a fair comparison with others, we set $N_O=100$ by default.

\noindent\textbf{Number of HCA Layers.}
We vary the number of HCA layers from 1 to 3 and report the results in~\cref{table:iici}. It shows that using two layers achieves the best performance on S3DIS.

\noindent\textbf{Momentum Coefficient.}
The momentum coefficient $\mu$ controls the evolving rate of our base prototypes. 
We explore its effects on performance in~\cref{table:mom}. The results show that varying $\mu$ causes minimal differences in mIoU, demonstrating the robustness of our proposed BPC module.

\section{Conclusion}
In this paper, we identify two critical issues in FS-PCS: foreground leakage and sparse point distribution, which have undermined the validity of previous progress and hindered further advancements.
To rectify these issues, we standardize FS-PCS by introducing a rigorous setting along with a new benchmark.
Moreover, we propose a novel correlation optimization paradigm that operates on CMC, diverging from the traditional feature optimization approach used by all previous FS-PCS models.
Building on this paradigm, our model~\ourmodel~incorporates HCA for effective contextual learning and BPC for background correlation adjustment, achieving state-of-the-art results across all FS-PCS settings. 
We hope that our work could serve as the foundation for FS-PCS and shed light on future research.

\myPara{Acknowledgments.} This work is supported in part by the Agency for Science, Technology and Research (A*STAR) under its MTC Young Individual Research Grant (Grant No. M21K3c0130), the Alexander von Humboldt Foundation, and the Pioneer Centre for AI, DNRF grant number P1.

{
    \small
    \bibliographystyle{ieeenat_fullname}
    \bibliography{main}
}

\section*{Appendix}

\begin{figure*}[!t]
    \centering
    \includegraphics[width=0.95\linewidth]{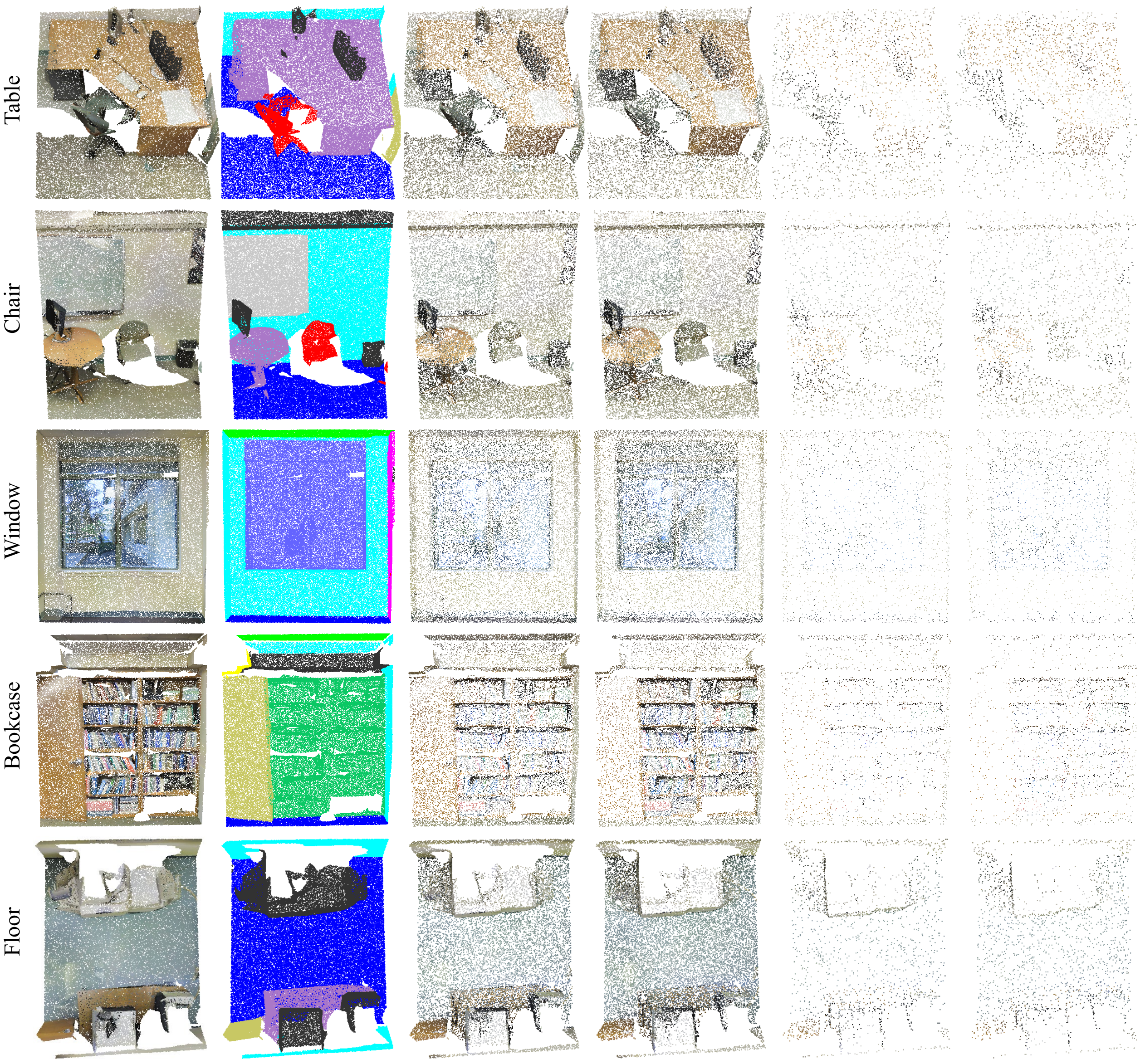}
    \caption{\textbf{Visualization of various scenes from the S3DIS dataset~\cite{armeni20163d}, with the target class for the $1$-way few-shot task labeled at the leftmost of each scene.} Each scene includes six types of point clouds, arranged \textit{from left to right:}
    (1) The original point cloud;
    (2) Ground truth of all categories;
    (3) Our corrected input with 20,480 points in a uniform distribution;
    (4) Input with 20,480 points in a biased distribution;
    (5) Input with 2,048 points in a uniform distribution;
    (6) Input with 2,048 points in a biased distribution, as adopted by previous works.}
    \label{fig:sampling-supp}
\end{figure*}

\subsection*{A. More Details about Foreground Leakage}
As discussed in~\cref{sec:pbs}, the current few-shot 3D point cloud semantic segmentation (FS-PCS) setting~\cite{zhao2021few,he2023prototype,ning2023boosting,zhu2023cross,mao2022bidirectional,wang2023few,zhang2023few} employs a non-uniform sampling mechanism with a bias toward foreground classes. This biased sampling algorithm samples more points from foreground objects than from the background, resulting in a noticeable point density disparity between foreground and background.

More precisely, the biased sampling algorithm can be outlined in Alg.~\ref{alg:smp}\footnote{The corresponding source code can be found at the~\href{https://github.com/Na-Z/attMPTI/blob/6e50296721475d917480fb3276b0ad81047f15e4/dataloaders/loader.py\#L39-L58}{link}.}. 
In line 1, it firstly obtains the input foreground point set $\matdn{P}{FG}$ that includes all the input points belonging to the foreground class $C$ with respect to the current few-shot task. 
Then, from lines 2 to 6, it calculates the quantity $N_{\rm FG}$ that will be used for sampling foreground points in the output. 
$N_{\rm FG}$ maintains a proportional relationship to the presence of foreground points in the input data when $n \geq m$. 
Next, in line 7, it selects $N_{\rm FG}$ points exclusively from the input foreground point set $\matdn{P}{FG}$. 
However, in line 8, the remaining $m-N_{\rm FG}$ points are sampled from the entire input points $\mathbf{X} = \{\matdn{P}{1},..,\matdn{P}{n}\}$, which still includes the foreground points in $\matdn{P}{FG}$.
Consequently, this double-sampling of foreground points in these two steps leads to foreground objects having a denser distribution of points in the final output than their background counterparts.

\begin{algorithm}
    \caption{The biased sampling algorithm}\label{alg:smp}
    \KwData{input point cloud $\mathbf{X}$ with $n$ points $\{\matdn{P}{1},..,\matdn{P}{n}\}$, sampling number $m$, foreground class $C$ with respect to current few-shot task}
    \KwResult{sampled points $\{\matdn{P}{i_1},..,\matdn{P}{i_m}\}$ from $\mathbf{X}$}
    $\matdn{P}{FG} \gets \{\matdn{P}{i}~|~label\_of(\matdn{P}{i}) = C\}$\;
      \eIf{$n < m$}{
        $N_{\rm FG} \gets |\matdn{P}{FG}|$\;
      }{
        $N_{\rm FG} \gets m \frac{|\matdn{P}{FG}|}{n}$\;
        }
    $\matdn{Res}{1} \gets$ sample $N_{\rm FG}$ points from $\matdn{P}{FG}$\;
    $\matdn{Res}{2} \gets$ sample $m - N_{\rm FG}$ points from $\mathbf{X}$\;
    $\{\matdn{P}{i_1},..,\matdn{P}{i_m}\} \gets \matdn{Res}{1} \cup \matdn{Res}{2}$\;
\end{algorithm}

We also present additional visualizations in~\cref{fig:sampling-supp}. Both the theoretical analysis and visualizations clearly demonstrate that this biased sampling leaks foreground class information to models through density disparity. 
Consequently, the models no longer need to excel at learning essential knowledge adaptation patterns for few-shot tasks; instead, they can simply segment the target by detecting denser regions. This foreground leakage undermines the validity of existing benchmarks of previous models.

\subsection*{B. More Implementation Details}
We employ the first three blocks from the Stratified Transformer~\cite{lai2022stratified} as our backbone. Our backbone architecture aligns with the one used for the S3DIS dataset~\cite{armeni20163d} in~\cite{lai2022stratified}, indicating that we maintain consistency in backbone architectures for both S3DIS and ScanNet~\cite{dai2017scannet}. Unlike~\cite{lai2022stratified}, we do not employ different Stratified Transformer architectures for these two datasets. The momentum coefficient $\mu$ within the BPC module is set to 0.995.
For both datasets, our input features include both the XYZ coordinates and RGB colors. The training and testing are using 4 RTX 3090 GPUs.

\begin{figure*}[t!]
    \centering
    \includegraphics[width=.98\linewidth]{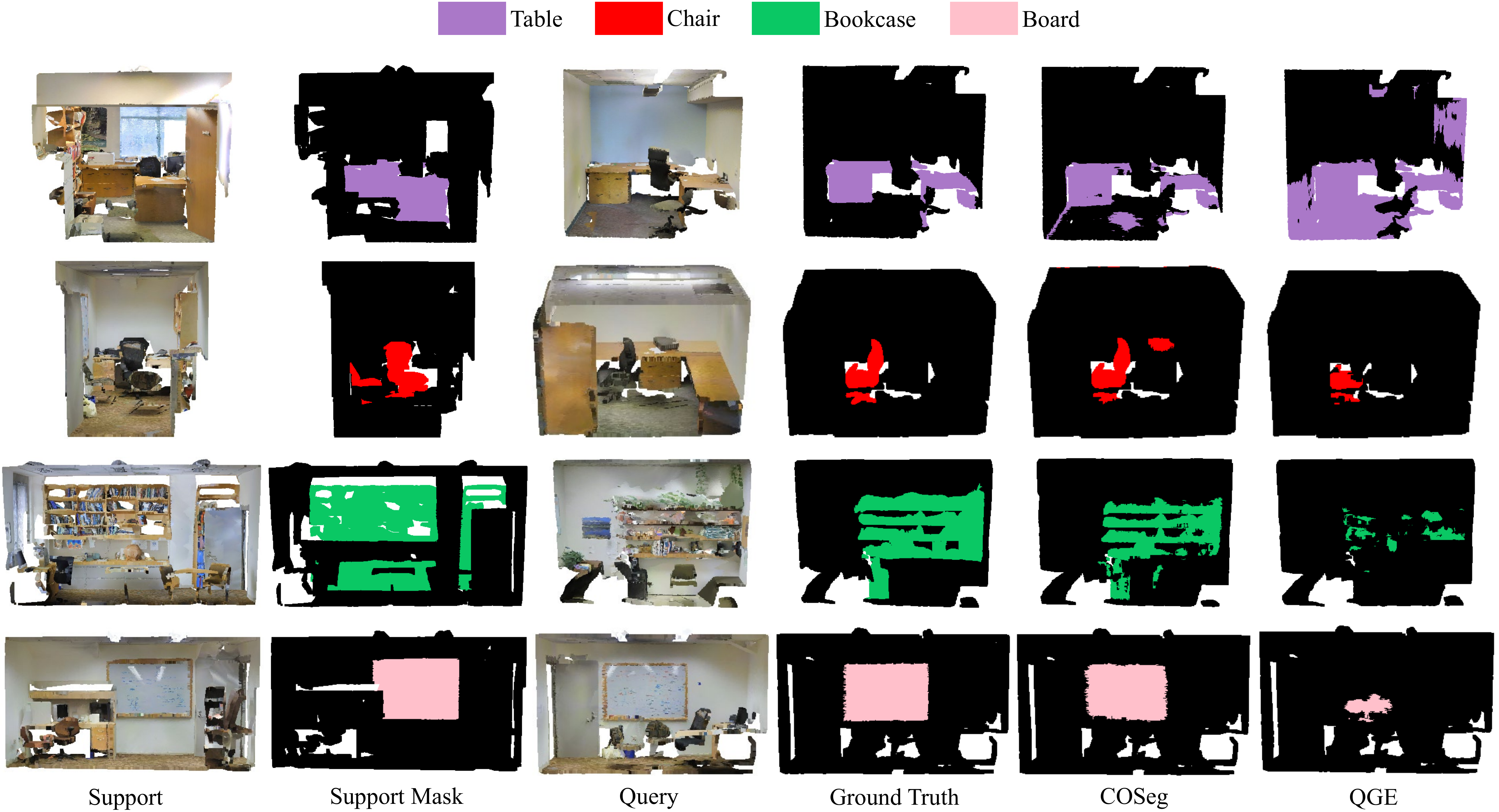}
    \caption{
    Qualitative comparisons between our proposed model~\ourmodel~and QGE~\cite{ning2023boosting}. 
    Each row, from top to bottom, represents the $1$-way $1$-shot task with the target category as table (\textcolor{mypurple}{purple}), chair (\textcolor{chair}{red}), bookcase (\textcolor{bookcase}{green}) and board (\textcolor{board}{pink}), respectively.
    }
    \label{fig:vs1-supp}
\end{figure*}

\begin{figure*}[t!]
    \vspace{-0.05in}
    \centering
    \includegraphics[width=.98\linewidth]{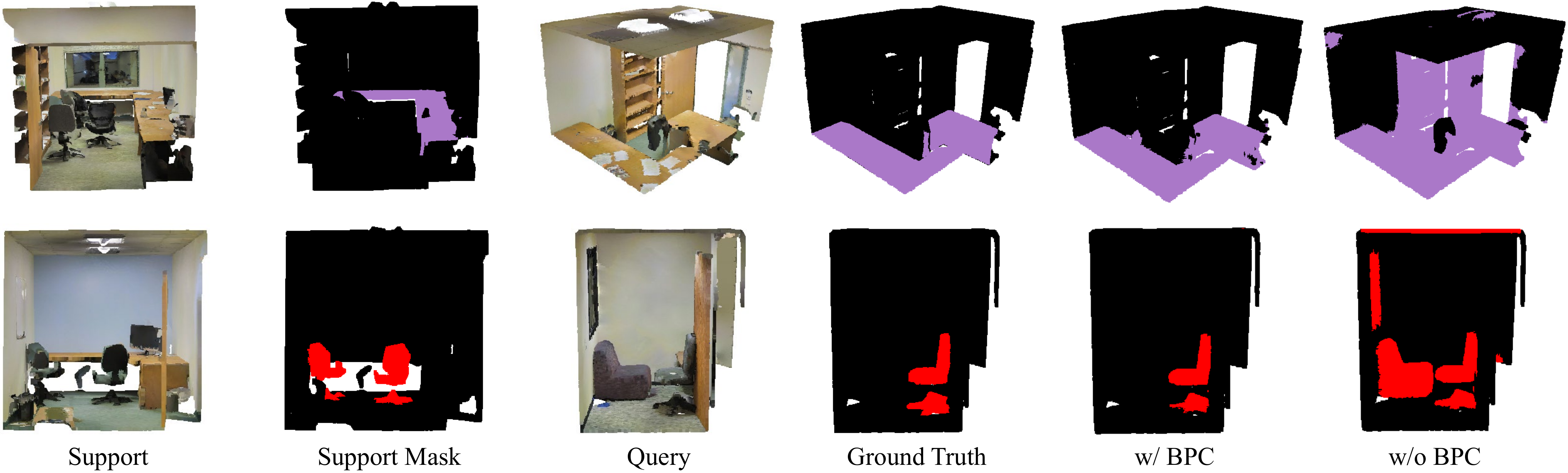}
    \caption{
    Qualitative comparisons between our models with BPC (w/ BPC) and without BPC (w/o BPC). 
    Each row has the target class under the $1$-way $1$-shot task as table (\textcolor{mypurple}{purple}) and chair (\textcolor{chair}{red}), respectively, arranged from top to bottom.
    }
    \label{fig:vs2-supp}
\end{figure*}

\subsection*{C. More Qualitative Results}
We present additional qualitative results in~\cref{fig:vs1-supp}, comparing our method ($5$th column) with the previous best-performing method, QGE ($6$th column). Besides,~\cref{fig:vs2-supp} showcases more visual comparisons between our models with BPC (w/ BPC, $5$th column) and without BPC (w/o BPC, $6$th column).

We have the following observations from the visual comparisons: (1) Our method yields visually better results than the previous best-performing method, highlighting the superiority of our proposed correlation optimization paradigm in enhancing the generalization ability for few-shot tasks. (2) The lightweight BPC module, equipped with non-parametric base prototypes, effectively mitigates the base susceptibility issue inherent in models. This ensures accurate segmentation of novel classes, further validating the efficacy of our approach.

\end{document}